\documentclass[sigconf]{acmart}
\AtBeginDocument{%
  \providecommand\BibTeX{{%
    \normalfont B\kern-0.5em{\scshape i\kern-0.25em b}\kern-0.8em\TeX}}}

\copyrightyear{2021}
\acmYear{2021}
\setcopyright{iw3c2w3}
\acmConference[WWW '21]{Proceedings of the Web Conference 2021}{April 19--23, 2021}{Ljubljana, Slovenia}
\acmBooktitle{Proceedings of the Web Conference 2021 (WWW '21), April 19--23, 2021, Ljubljana, Slovenia}
\acmPrice{}
\acmDOI{10.1145/3442381.3450023}
\acmISBN{978-1-4503-8312-7/21/04}

\usepackage{amsmath,amsfonts,amsthm}
\usepackage[ruled,vlined,shortend, linesnumbered]{algorithm2e} 
\usepackage{algpseudocode}
\usepackage{textcomp}
\usepackage{subcaption}
\usepackage{times}
\usepackage{soul}
\usepackage{graphicx}
\usepackage{hyperref}

\usepackage{bbding} 
\usepackage{xcolor}
\usepackage{color, colortbl}
\usepackage{enumitem}
\usepackage{amsfonts}
\usepackage{helvet}
\usepackage{courier}
\usepackage{multirow}
\usepackage{booktabs}
\usepackage{placeins}
\newsavebox{\measurebox}
\newtheorem{definition}{Definition}

\newcommand{\method}{\textsc{MStream}}
\newcommand{\densealert}{\textsc{DenseAlert}}
\newcommand{\rcf}{Random Cut Forest}
\newcommand{\iso}{Isolation Forest}
\newcommand{\elliptic}{Elliptic Envelope}
\newcommand{\lof}{Local Outlier Factor}
\renewcommand{\vec}[1]{\mathbf{#1}}

\begin{document}

\title{\method: Fast Anomaly Detection in Multi-Aspect Streams}

\author{Siddharth Bhatia}

\affiliation{%
  \institution{National University of Singapore}
}
\email{siddharth@comp.nus.edu.sg}

\author{Arjit Jain}
\affiliation{%
  \institution{IIT Bombay}
}
\email{arjit@cse.iitb.ac.in}

\author{Pan Li}
\affiliation{%
  \institution{Purdue University}
}
\email{panli@purdue.edu}

\author{Ritesh Kumar}
\affiliation{%
  \institution{IIT Kanpur}
}
\email{riteshk@iitk.ac.in}

\author{Bryan Hooi}
\affiliation{%
  \institution{National University of Singapore}
}
\email{bhooi@comp.nus.edu.sg}


\begin{abstract}
  Given a stream of entries in a multi-aspect data setting i.e., entries having multiple dimensions, how can we detect anomalous activities in an unsupervised manner? For example, in the intrusion detection setting, existing work seeks to detect anomalous events or edges in dynamic graph streams, but this does not allow us to take into account additional attributes of each entry. Our work aims to define a streaming multi-aspect data anomaly detection framework, termed \method\, which can detect unusual group anomalies as they occur, in a dynamic manner. \method\ has the following properties: (a) it detects anomalies in multi-aspect data including both categorical and numeric attributes; (b) it is online, thus processing each record in constant time and constant memory; (c) it can capture the correlation between multiple aspects of the data. \method\ is evaluated over the \emph{KDDCUP99}, \emph{CICIDS-DoS}, \emph{UNSW-NB 15} and \emph{CICIDS-DDoS} datasets, and outperforms state-of-the-art baselines.
\end{abstract}



\maketitle

\section{Introduction}

Given a stream of entries (i.e. records) in \emph{multi-aspect data} (i.e. data having multiple features or dimensions), how can we detect anomalous behavior, including group anomalies involving the sudden appearance of large groups of suspicious activity, in an unsupervised manner?

In particular, we focus on an important application of anomaly detection to intrusion detection in networks. In this application, we want to design algorithms that monitor a stream of records, each of which represents a single connection (or `flow') over the network.
We aim to detect multiple types of suspicious activities, such as denial of service or port scanning attacks, whereby attackers make a large number of connections to a target server to make it inaccessible or to look for vulnerabilities. 

Recent intrusion detection datasets typically report tens of features for each individual flow, such as its source and destination IP, port, protocol, average packet size, etc. This makes it important to design approaches that can handle \emph{multi-aspect data}. In addition, to effectively guard against attacks, it is important for our algorithm to process the data in a \emph{streaming} manner, so that we can quickly report any attack in real-time, as soon as it arrives.

Some existing approaches for this problem aim to detect \emph{point anomalies}, or individually unusual connections. However, as this ignores the relationships between records, it does not effectively detect large and suddenly appearing \emph{groups} of connections, as is the case in denial of service and other attacks. For detecting such groups, there are also existing methods based on dense subgraph detection~\cite{bhatia2020midas} as well as dense subtensor detection~\cite{shin2017densealert,sun2006beyond}. However, these approaches are generally designed for datasets with a smaller number of dimensions, thus facing significant difficulties scaling to our dataset sizes. Moreover, they treat all variables of the dataset as categorical variables, whereas our approach can handle arbitrary mixtures of categorical variables (e.g. source IP address) and numerical variables (e.g. average packet size). 

Hence, in this work, we propose \method, a method for processing a stream of multi-aspect data that detects \emph{group anomalies}, i.e. the sudden appearance of large amounts of suspiciously similar activity. Our approach naturally allows for similarity both in terms of categorical variables (e.g. a small group of repeated IP addresses creating a large number of connections), as well as in numerical variables (e.g. numerically similar values for average packet size).

\method\ is a streaming approach that performs each update in constant memory and time. This is constant both with respect to the stream length as well as in the number of attribute values for each attribute: this contrasts with tensor decomposition-based approaches such as STA and dense subtensor-based approaches such as \densealert, where memory usage grows in the number of possible attribute values. To do this, our approach makes use of locality-sensitive hash functions (LSH), which process the data in a streaming manner while allowing connections which form group anomalies to be jointly detected, as they consist of similar attribute values and hence are mapped into similar buckets by the hash functions. Finally, we demonstrate that the anomalies detected by \method\ are explainable.

To incorporate correlation between features, we further propose \method-PCA, \method-IB and \method-AE which leverage Principal Component Analysis (PCA), Information Bottleneck (IB), and Autoencoders (AE) respectively, to map the original features into a lower-dimensional space and then execute \method\ in this lower-dimensional space. \method-AE is shown to provide better anomaly detection performance while also improving speed compared to \method, due to its lower number of dimensions.

In summary, the main contributions of our approach are:
\begin{enumerate}
    \item {\bf Multi-Aspect Group Anomaly Detection:} We propose a novel approach for detecting group anomalies in multi-aspect data, including both categorical and numeric attributes.
    \item {\bf Streaming Approach:} Our approach processes the data in a fast and streaming fashion, performing each update in constant time and memory.
    \item {\bf Effectiveness:} Our experimental results show that \method\ outperforms baseline approaches.
\end{enumerate}
{\bf Reproducibility}: Our code and datasets are publicly available at \href{https://github.com/Stream-AD/MStream}{https://github.com/Stream-AD/MStream}.

\section{Related Work}
\label{sec:rel}

\begin{table*}[!htb]
\centering
\caption{Comparison of relevant multi-aspect anomaly detection approaches.}
\label{tab:comparison}
\begin{tabular}{@{}rcccccccc|c@{}}
\toprule
& {Elliptic }
& {LOF }
& {I-Forest }
& {STA }
& {MASTA }
& {STenSr }
& {\rcf} 
& {\densealert } 
& {\bf {\method}} \\ 
& ($1999$) & ($2000$) & ($2008$) & ($2006$) & ($2015$) & ($2015$) & ($2016$) & ($2017$) & ($2021$) \\\midrule
\textbf{Group Anomalies} & & & & & & & & \Checkmark & \CheckmarkBold \\
\textbf{Real-valued Features} & \Checkmark & \Checkmark & \Checkmark & & & & \Checkmark & & \CheckmarkBold \\
\textbf{Constant Memory} & & & & & & & \Checkmark & \Checkmark & \CheckmarkBold \\
\textbf{Const. Update Time} & & & & \Checkmark & \Checkmark & \Checkmark & \Checkmark & \Checkmark & \CheckmarkBold \\
\bottomrule
\end{tabular}
\end{table*}

Our work is closely related to areas like anomaly detection on graphs \cite{akoglu2015graph, DBLP:conf/pakdd/ZhangLYFC19, Malliaros2012FastRE, DBLP:conf/sdm/BogdanovFMPRS13, 10.1145/2213836.2213974, Gupta2017LookOutOT, 7836684, 10.1145/3178876.3186056, 10.1145/3139241, 10.1145/2939672.2939734, perozzi2016scalable, DBLP:journals/wias/BonchiBGS19, 7817049,tong2011non, yoon2019fast}, graph and stream classification \cite{8016599, 10.1007/978-3-642-13657-3_52, 6144793, 6884853, 6544842, wangprovably}, and outlier detection on streams \cite{10.1145/3219819.3220022, WILMET2019197, DBLP:journals/corr/abs-1906-02524, DBLP:conf/kdd/ManzoorLA18, 7837870, sun2019fast}. In this section, we limit our review only to previous approaches for detecting anomalies on edge-streams, tensors and multi-aspect data. See \cite{fanaee2016tensor} for an extensive survey on tensor-based anomaly detection.

\noindent{\bf Anomaly Detection in Edge Streams} uses as input a stream of edges over time. We categorize them according to the type of anomaly detected.
\begin{itemize}
\item Anomalous node detection:
Given an edge stream, \cite{yu2013anomalous} detects nodes whose egonets suddenly and significantly change.
\item Anomalous subgraph detection:
Given an edge stream, \textsc{DenseAlert} \cite{shin2017densealert} identifies dense subtensors created within a short time.
\item Anomalous edge detection:
RHSS \cite{ranshous2016scalable} focuses on sparsely-connected parts of a graph, while \textsc{SedanSpot} \cite{eswaran2018sedanspot} identifies edge anomalies based on edge occurrence, preferential attachment, and mutual neighbors. \textsc{Midas} \cite{bhatia2020midas} identifies microcluster based anomalies, or suddenly arriving groups of suspiciously similar edges.
\end{itemize}

\noindent{\bf Anomaly Detection in Multi-Aspect Data Streams} uses as input a stream of multi-aspect data records over time. Each multi-aspect data record can also be considered as an edge of an attributed graph having multiple attributes. Therefore, in addition to detecting anomalies in multi-aspect data streams, the following approaches can also detect anomalies in edge streams.

\begin{itemize}
\item Score Plot based: Score Plots are obtained from tensor decomposition which are then analyzed manually or automatically for anomaly detection. These score plots can be one-dimensional: \cite{papalexakis2014spotting}, multi-dimensional: \textsc{MalSpot} \cite{mao2014malspot} or time-series  \cite{papalexakis2012parcube}.

\item Histogram based: MASTA \cite{fanaee2015multi} uses histogram approximation to analyze tensors. It vectorizes the whole tensor and simultaneously segments into slices in each mode.
The distribution of each slice is compared against the vectorized tensor to identify anomalous slices.

\item Tensor decomposition based:
Tensor decomposition methods such as \cite{kolda2009tensor} can be used to find anomalies. \cite{zhou2016accelerating} and STA \cite{sun2006beyond} are streaming algorithms for CPD and Tucker decompositions.  STenSr \cite{shi2015stensr} models the tensor stream as a single incremental tensor for representing the entire network, instead of dealing with each tensor in the stream separately. \cite{li2011robust} uses subspace learning in tensors to find anomalies. STA monitors the streaming decomposition reconstruction error for each tensor at each time instant and anomalies occur when this error goes beyond a pre-defined threshold. However \cite{shin2017densealert} shows limited accuracy for dense-subtensor detection based on tensor decomposition.

\item Dense subtensor detection based:
Dense-subtensor detection has been used to detect anomalies in \textsc{M-Zoom} \cite{shin2016m}, \textsc{D-Cube} \cite{shin2017d}, \cite{maruhashi2011multiaspectforensics} and \textsc{CrossSpot} \cite{jiang2015general} but these approaches consider the data as a static tensor. \textsc{DenseAlert} \cite{shin2017densealert} is a streaming algorithm to identify dense subtensors created within a short time.

\end{itemize}

\noindent{\textbf{Other Approaches for Anomaly Detection}} can typically be used in multi-aspect settings by converting categorical attributes to numerical ones e.g. using one-hot encoding. Elliptic Envelope \cite{rousseeuw1999fast} fits an ellipse to the normal data points by fitting a robust covariance estimate to the data. Local Outlier Factor (LOF) \cite{breunig2000lof} estimates the local density at each point, then identifies anomalies as points with much lower local density than their neighbors. Isolation Forest (I-Forest) \cite{liu2008isolation} constructs trees by randomly selecting features and splitting them at random split points, and then defines anomalies as points which are separated from the rest of the data at low depth values. Random Cut Forest (RCF) \cite{guha2016robust} improves upon isolation forest by creating multiple random cuts (trees) of data and constructs a forest of such trees to determine whether a point is anomalous or not.

Recently, deep learning approaches for anomaly detection in multi-aspect data have also been proposed. DAGMM \cite{zong2018deep} learns a Gaussian Mixture density model (GMM) over a low-dimensional latent space produced by a deep autoencoder. \cite{JU2020167} use metric learning for anomaly detection. Deep Structured Energy-based Model for Anomaly Detection (DSEBM) \cite{zhai2016deep} trains deep energy models such as Convolutional and Recurrent EBMs using denoising score matching instead of maximum likelihood, for performing anomaly detection. More recently, methods like APAE \cite{Goodge2020RobustnessOA}, MEG \cite{kumar2019maximum} and Fence GAN \cite{ngo2019} have been successfully used to detect anomalies. 

For the task of Intrusion Detection \cite{Gradison2018, 1565245, 10.1145/3143361.3143399, 10.1016/j.cose.2006.05.005, WANG2014103, 4400777}, a variety of different approaches have been used in the literature including Ensemble methods \cite{DBLP:journals/scn/RajagopalKH20}, Feature Selection \cite{c5d37e2294e349d59233c4b6e41cae3a}, Fuzzy Neural Networks \cite{fuzzynn}, Kernel Methods \cite{10.1007/11427469_77}, Random Forests \cite{8766544}, and deep learning based methods \cite{8681044} \cite{8643036}. However, we refrain from comparing with these approaches as they do not process the data in a streaming manner and typically require large amount of labelled training data, whereas we process the data in an unsupervised and online manner. 

Note that \lof, \iso, \elliptic, STA, MASTA, STenSr, \densealert\ and \rcf\ are unsupervised algorithms. Of these, only \densealert\ performs group anomaly detection (by detecting dense subtensors); however, as shown in Table \ref{tab:comparison}, it cannot effectively handle real-valued features (as it treats all features as discrete-valued).

\section{Problem}

Let $\mathcal{R} = \{r_1, r_2, \hdots\}$ be a stream of records, arriving in a streaming manner. Each record $r_i = (r_{i1}, \hdots, r_{id})$ consists of $d$ \emph{attributes} or dimensions, in which each dimension can either be categorical (e.g. IP address) or real-valued (e.g. average packet length). Note that since the data is arriving over time as a stream, we do not assume that the set of possible feature values is known beforehand; for example, in network traffic settings, it is common for new IP addresses to be seen for the first time at some point in the middle of the stream.

Our goal is to detect \emph{group anomalies}. Intuitively, group anomalies should have the following properties:

\begin{enumerate}
    \item {\bf Similarity in Categorical Attributes:} for categorical attributes, the group anomalies consist of a relatively small number of attribute values, repeated a suspiciously large number of times.
    \item {\bf Similarity in Real-Valued Attributes:} for real-valued attributes, the group anomalies consist of clusters of numerically similar attribute values.
    \item {\bf Temporally Sudden:} the group anomalies arrive suddenly, over a suspiciously short amount of time. In addition, their behavior (in terms of attribute values) should clearly differ from what we have observed previously, over the course of the stream.
\end{enumerate}
\section{Proposed Algorithm}

\subsection{Motivation}

Consider the toy example in Table \ref{tab:toy}, comprising a stream of connections over time. This dataset shows a clear block of suspicious activity from time $4$ to $5$, consisting of several IP addresses repeated a large number of times, as well as large packet sizes which seem to be anomalously large compared to the usual distribution of packet sizes.
\begin{table}[!htb]
\centering
\caption{Simple toy example, consisting of a stream of multi-aspect connections over time.}
\label{tab:toy}
\addtolength{\tabcolsep}{-2pt}
\begin{tabular}{@{}ccccc@{}}
\toprule
{\bf Time} & {\bf Source IP} & {\bf Dest. IP} & {\bf Pkt. Size} & {\bf $\cdots$} \\ \midrule
$1$ & \ \ \ $194.027.251.021$ \ \ \ & $194.027.251.021$ & $100$ & $\cdots$ \\
$2$ & \ \ \ $172.016.113.105$ \ \ \ & $207.230.054.203$ & $80$ & $\cdots$ \\
\textcolor{red}{$4$} & \ \ \ \textcolor{red}{$194.027.251.021$} \ \ \ & \textcolor{red}{$192.168.001.001$} & \textcolor{red}{$1000$} & \textcolor{red}{$\cdots$} \\
\textcolor{red}{$4$} & \ \ \ \textcolor{red}{$194.027.251.021$} \ \ \ & \textcolor{red}{$192.168.001.001$} & \textcolor{red}{$995$} & \textcolor{red}{$\cdots$} \\
\textcolor{red}{$4$} & \ \ \ \textcolor{red}{$194.027.251.021$} \ \ \ & \textcolor{red}{$192.168.001.001$} & \textcolor{red}{$1000$} & \textcolor{red}{$\cdots$} \\
\textcolor{red}{$5$} & \ \ \ \textcolor{red}{$194.027.251.021$} \ \ \ & \textcolor{red}{$192.168.001.001$} & \textcolor{red}{$990$} & \textcolor{red}{$\cdots$} \\
\textcolor{red}{$5$} & \ \ \ \textcolor{red}{$194.027.251.021$} \ \ \ & \textcolor{red}{$194.027.251.021$} & \textcolor{red}{$1000$} & \textcolor{red}{$\cdots$} \\
\textcolor{red}{$5$} & \ \ \ \textcolor{red}{$194.027.251.021$} \ \ \ & \textcolor{red}{$194.027.251.021$} & \textcolor{red}{$995$} & \textcolor{red}{$\cdots$} \\
$6$ & \ \ \ $194.027.251.021$ \ \ \ & $194.027.251.021$ & $100$ & $\cdots$ \\
$7$ & \ \ \ $172.016.113.105$ \ \ \ & $207.230.054.203$ & $80$ & $\cdots$ \\
\bottomrule
\end{tabular}
\end{table}

The main challenge, however, is to detect this type of patterns in a {\bf streaming} manner, considering that we do not want to set any limits a priori on the duration of the anomalous activity we want to detect, or the number of IP addresses (or other attribute values) which may be involved in this activity.

As shown in Figure \ref{fig:mstream}, our approach addresses these problems through the use of a number of {\bf locality-sensitive hash functions}~\cite{charikar2002similarity}
which hash each incoming tuple into a fixed number of buckets. Intuitively, we do this such that tuples with many similar entries tend to be hashed into similar buckets. These hash functions are combined with a {\bf temporal scoring} approach, which takes into account how much overlap we observe between the buckets at any time: high amounts of overlap arriving in a short period of time suggest the presence of anomalous activity.

In Sections \ref{sec:hashfunc} and \ref{sec:scoring}, we describe our \method\ approach, and in Section \ref{sec:mstreamae}, we describe our \method-PCA, \method-IB and \method-AE approaches which incorporate correlation between features in an unsupervised manner. \method-PCA uses principal component analysis, \method-IB uses information bottleneck, and \method-AE uses an autoencoder to first compress the original features and then apply \method\ in the compressed feature space.

\begin{figure*}[!htb]
        \center{\includegraphics[width=0.958\linewidth]
        {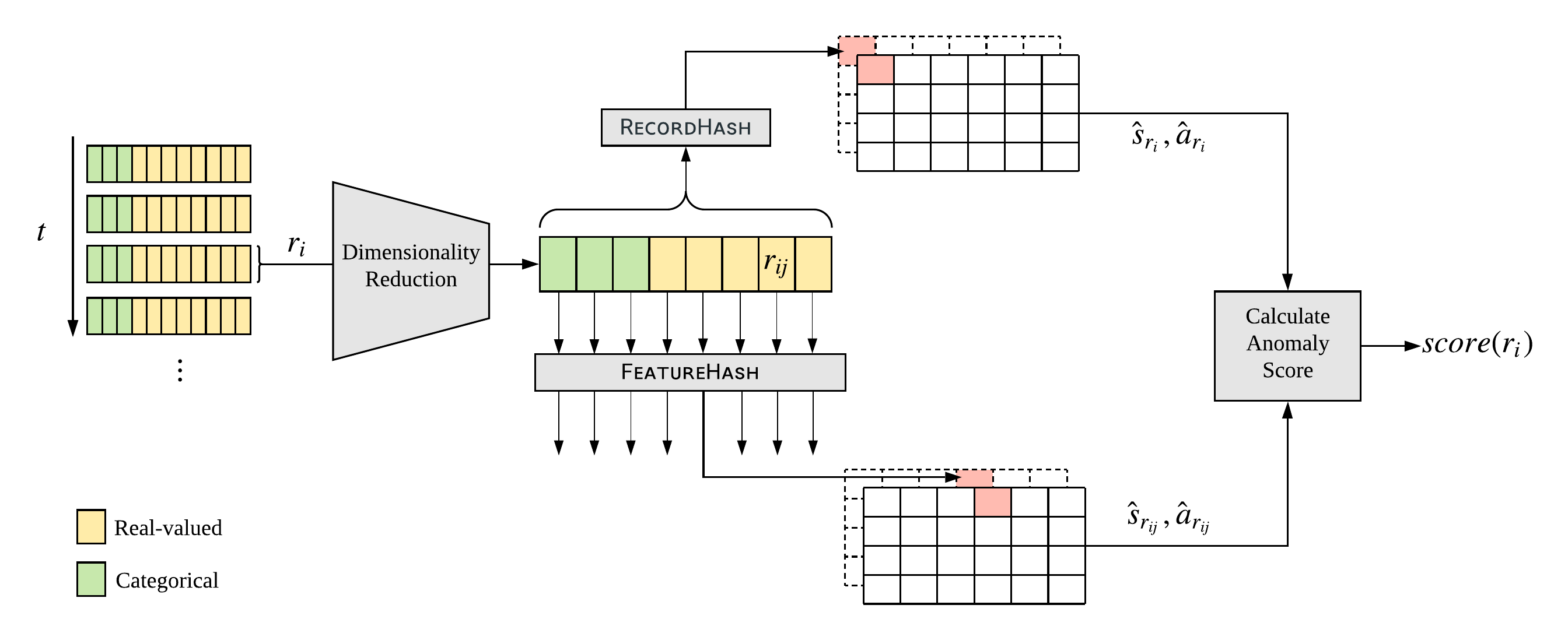}}
        \caption{\label{fig:mstream} Diagram of the proposed \method. The dimensionality reduction unit (Section \ref{sec:mstreamae}) takes in a record and outputs a lower-dimensional embedding. Two types of locality-sensitive hash functions are then applied. \textsc{FeatureHash} (Algorithm \ref{alg:featurehash}) hashes each individual feature and \textsc{RecordHash} (Algorithm \ref{alg:recordhash}) hashes the entire record jointly. These are then combined together using a temporal scoring approach to calculate the anomaly score for the record (Algorithm \ref{alg:mstream}).}
        \Description{\label{fig:mstream} Diagram of the proposed \method. The dimensionality reduction unit (Section \ref{sec:mstreamae}) takes in a record and outputs a lower-dimensional embedding. Two types of locality-sensitive hash functions are then applied. \textsc{FeatureHash} (Algorithm \ref{alg:featurehash}) hashes each individual feature and \textsc{RecordHash} (Algorithm \ref{alg:recordhash}) hashes the entire record jointly. These are then combined together using a temporal scoring approach to calculate the anomaly score for the record (Algorithm \ref{alg:mstream}).}
\end{figure*}

\subsection{Hash Functions}
\label{sec:hashfunc}

Our approach uses two types of hash functions: \textsc{FeatureHash}, which hashes each feature individually, and \textsc{RecordHash}, which hashes an entire record jointly. We use multiple independent copies of each type of hash function, and explain how to combine these to produce a single anomalousness score.

\subsubsection{FeatureHash}
\label{sec:feature}

As shown in Algorithm \ref{alg:featurehash}, \textsc{FeatureHash} consists of hash functions independently applied to a single feature. There are two cases, corresponding to whether the feature is categorical (e.g. IP address) or real-valued (e.g. average packet length):

For \textbf{categorical} data, we use standard linear hash functions \cite{litwin1980linear} which map integer-valued data randomly into $b$ buckets, i.e. $\{0, \dots, b-1\}$, where $b$ is a fixed number.

For \textbf{real-valued} data, however, we find that randomized hash functions tend to lead to highly uneven bucket distributions for certain input datasets. Instead, we use a streaming log-bucketization approach. We first apply a log-transform to the data value (line 5), then perform min-max normalization, where the min and max are maintained in a streaming manner (line 7) and finally map it such that the range of feature values is evenly divided into $b$ buckets, i.e. $\{0, \dots, b-1\}$ (line 8). 
    
\begin{algorithm}
	\caption{\textsc{FeatureHash}: Hashing Individual Feature  \label{alg:featurehash}}
	\KwIn{$r_{ij}$ (Feature $j$ of record $r_{i}$)}
	\KwOut{Bucket index in $\{0, \dots, b-1\}$ to map $r_{ij}$ into}
	\textbf{if} $r_{ij}$ is categorical  \\
	\ \ \ \ \textbf{output} $\Call{HASH}{r_{ij}}$ \tcp*[f]{Linear Hash~\cite{litwin1980linear}} \\
	\textbf{else if} $r_{ij}$ is real-valued  \\
	\ \ \ \ {\bf $\triangleright$ Log-Transform} \\
	\ \ \ \ \ \ \ \ $\tilde{r}_{ij} = \log(1+r_{ij})$ \\
	\ \ \ \ {\bf $\triangleright$ Normalize} \\
	\ \ \ \ \ \ \ \ $\tilde{r}_{ij} \gets \frac{\tilde{r}_{ij} - min_{j}}{max_{j} - min_{j}}$ \tcp*[f]{Streaming Min-Max} \\
    \ \ \ \ {\bf output} $\lfloor \tilde{r}_{ij} \cdot b \rfloor  ($mod$ \ b)$ \tcp*[f]{Bucketization into $b$ buckets} \\
\end{algorithm}

\subsubsection{RecordHash}
\label{sec:record}

As shown in Algorithm \ref{alg:recordhash}, in \textsc{RecordHash}, we operate on all features of a record simultaneously. We first divide the entire record $r_{i}$ into two parts, one consisting of the categorical features $\mathcal{C}$, say $r_{i}^{cat}$, and the other consisting of real-valued features $\mathcal{R}$, say $r_{i}^{num}$. We then separately hash $r_{i}^{cat}$ to get $bucket_{cat}$, and $r_{i}^{num}$ to get $bucket_{num}$. Finally we take the sum modulo $b$ of $bucket_{cat}$ and $bucket_{num}$ to get a bucket for $r_{i}$. We hash $r_{i}^{cat}$ and $r_{i}^{num}$ as follows:

\begin{enumerate}
\item $r_{i}^{cat}$: We use standard linear hash functions \cite{litwin1980linear} to map $\forall j \in \mathcal{C}$ each of the individual features $r_{ij}$ into $b$ buckets, and then combine them by summing them modulo $b$ to compute the bucket index $bucket_{cat}$ for $r_{i}^{cat}$ (line 3).

\item $r_{i}^{num}$: To compute the hash of a real-valued record $r_{i}^{num}$ of dimension $p=|\mathcal{R}|$, we choose $k$ random vectors $\vec{a_{1}},\vec{a_{2}},..,\vec{a_{k}}$ each having $p$ dimensions and independently sampled from a Gaussian distribution $\mathcal{N}_{p}(\vec{0},\vec{I_p})$, where $k=\lceil \log_{2}(b) \rceil$. We compute the scalar product of $r_{i}^{num}$ with each of these vectors (line 6). We then map the positive scalar products to $1$ and the non-positive scalar products to $0$ and then concatenate these mapped values to get a $k$-bit string, then convert it from a bitset into an integer $bucket_{num}$ between $0$ and $2^k-1$. (line 10).
\end{enumerate}

\begin{algorithm}
	\caption{\textsc{RecordHash}: Hashing Entire Record  \label{alg:recordhash}}
	\KwIn{Record $r_{i}$}
	\KwOut{Bucket index in $\{0, \dots, b-1\}$ to map $r_{i}$ into}
	{\bf $\triangleright$ Divide $r_{i}$ into its categorical part, $r_{i}^{cat}$, and its numerical part, $r_{i}^{num}$} \\
	{\bf $\triangleright$ Hashing $r_{i}^{cat}$} \\
	\ \ \ \ $bucket_{cat} = (\sum_{j \in \mathcal{C} }${$\Call{HASH}{r_{ij}})$} (mod $b$) \tcp*[f]{Linear Hash~\cite{litwin1980linear}} \\
	{\bf $\triangleright$ Hashing $r_{i}^{num}$} \\
	\ \ \ \ \textbf{for} $id \gets$ $1$ to $k$  \\
	\ \ \ \ \ \ \ \	\textbf{if} $\langle{r_{i}^{num} , \vec{a_{id}}}\rangle > 0$  \\
	\ \ \ \ \ \ \ \ \ \ \ \ $bitset[id] = 1$ \\
	\ \ \ \ \ \ \ \ \textbf{else}  \\
	\ \ \ \ \ \ \ \ \ \ \ \ $bitset[id] = 0$ \\
	\ \ \ \ $bucket_{num} = \Call{INT}{bitset}$ \tcp*[f]{Convert bitset to integer} \\

    {\bf output}\ $(bucket_{cat}+bucket_{num})($mod$ \ b)$
\end{algorithm}

\subsection{Temporal Scoring}
\label{sec:scoring}

A recent algorithm, \textsc{Midas} \cite{bhatia2020midas}, finds anomalous edges of a dynamic graph in a streaming manner. It combines a chi-squared goodness-of-fit test with the Count-Min-Sketch (CMS)~\cite{cormode2005improved} streaming data structures to get an anomaly score for each edge. In \textsc{Midas}, $s_{uv}$ is defined to be the total number of edges from node $u$ to $v$ up to the current time $t$, while $a_{uv}$ is the number of edges from node $u$ to $v$ in the current time tick $t$ (but not including past time ticks). \textsc{Midas} then divides the edges into two classes: edges at the current time tick $t$ ($=a_{uv}$), and edges in past time ticks ($=s_{uv} - a_{uv}$), and computes the chi-squared statistic i.e. the sum over categories of $\frac{(\text{observed} - \text{expected})^2}{\text{expected}}$:

\begin{align*}
X^2 &= \frac{(\text{observed}_{(cur\_t)} - \text{expected}_{(cur\_t)})^2}{\text{expected}_{(cur\_t)}} \\
&+ \frac{(\text{observed}_{(past\_t)} - \text{expected}_{(past\_t)})^2}{\text{expected}_{(past\_t)}}\\
\end{align*}
\begin{align*}
&= \frac{(a_{uv} - \frac{s_{uv}}{t})^2}{\frac{s_{uv}}{t}} + \frac{((s_{uv} - a_{uv}) - \frac{t-1}{t} s_{uv})^2}{\frac{t-1}{t} s_{uv}}\\
&= \left(a_{uv} - \frac{s_{uv}}{t}\right)^2 \frac{t^2}{s_{uv}(t-1)}
\end{align*}

\textsc{Midas} uses two types of CMS data structures to maintain approximate counts $\hat{s}_{uv}$ and $\hat{a}_{uv}$ which estimate $s_{uv}$ and $a_{uv}$ respectively. The anomaly score for an edge in \textsc{Midas} is then defined as:

\begin{equation}
score(u,v,t) = \left(\hat{a}_{uv} - \frac{\hat{s}_{uv}}{t}\right)^2 \label{eqn:eq1} \frac{t^2}{\hat{s}_{uv}(t-1)}
\end{equation}

However, \textsc{Midas} is designed to detect anomalous edges, which are two-dimensional records (consisting of source and destination node index).
Therefore, it cannot be applied in the high-dimensional setting of multi-aspect data. Moreover, \textsc{Midas} treats variables of the dataset as categorical variables, whereas multi-aspect data can contain arbitrary mixtures of categorical variables (e.g. source IP address) and numerical variables (e.g. average packet size).

We extend \textsc{Midas} to define an anomalousness score for each record and detect anomalous records in a streaming manner. Given each incoming record $r_{i}$ having $j$ features, we can compute $j+1$ anomalousness scores: one for the entire record $r_{i}$ and one for each individual feature $r_{ij}$. We compute each score by computing the chi-squared statistic over the two categories: current time tick and past time ticks. Anomaly scores for individual attributes are useful for interpretability, as they help explain which features are most responsible for the anomalousness of the record.
Finally, we combine these scores by taking their sum.

\begin{definition}[Anomaly Score]
Given a newly arriving record $(r_{i},t)$, our anomalousness score is computed as:
\begin{align}
score (r_{i},t) = \left(\hat{a}_{r_{i}} - \frac{\hat{s}_{r_{i}}}{t}\right)^2 \frac{t^2}{\hat{s}_{r_{i}}(t-1)} + \sum_{j=1}^{d} score(r_{ij},t)
\end{align}
where,
\begin{align}
\text{score}(r_{ij},t) = \left(\hat{a}_{r_{ij}} - \frac{\hat{s}_{r_{ij}}}{t}\right)^2 \frac{t^2}{\hat{s}_{r_{ij}}(t-1)}
\end{align}
\end{definition}
and $\hat{a}_{r_{i}} ($or $\hat{a}_{r_{ij}})$ is an approximate count of  $r_{i} ($or $r_{ij})$ at current time $t$ and $\hat{s}_{r_{i}} ($or $\hat{s}_{r_{ij}})$ is an approximate count of $r_{i} ($or $r_{ij})$ up to time $t$.

We also allow temporal flexibility of records, i.e. records in the recent past count towards the current anomalousness score. This is achieved by reducing the counts $\hat{a}_{r_{i}}$ and $\hat{a}_{r_{ij}} \forall j \in \{1,..,d\}$ by a factor of $\alpha \in (0, 1)$ rather than resetting them at the end of each time tick. This results in past records counting towards the current time tick, with a diminishing weight.

\method\ is summarised in Algorithm \ref{alg:mstream}.

\begin{algorithm}
	\caption{\method:\ Streaming Anomaly Scoring \label{alg:mstream}}
	\KwIn{Stream of records over time}
	\KwOut{Anomaly scores for each record}
	{\bf $\triangleright$ Initialize data structures:} \\
	\ \ \ \ Total record count $\hat{s}_{r_{i}}$ and total attribute count $\hat{s}_{r_{ij}} \forall j \in \{1,..,d\}$ \\
	\ \ \ \ Current record count $\hat{a}_{r_{i}}$ and current attribute count  $\hat{a}_{r_{ij}} \forall j \in \{1,..,d\}$ \\
	\While{new record $(r_i,t) = (r_{i1}, \hdots, r_{id},t)$ is received:}{
	{\bf $\triangleright$ Hash and Update Counts:} \\

	\ \ \ \ \textbf{for} $j \gets$ $1$ to $d$  \\
	\ \ \ \ \ \ \ \ $bucket_{j} = \Call{FeatureHash}{r_{ij}}$ \\
    \ \ \ \ \ \ \ \ Update count of $bucket_{j}$ \\
	
	\ \ \ \ $bucket= \Call{RecordHash}{r_{i}}$\\
	\ \ \ \ Update count of $bucket$\\
	{\bf $\triangleright$ Query Counts:} \\
	\ \ \ \ Retrieve updated counts $\hat{s}_{r_{i}}$, $\hat{a}_{r_{i}}$, $\hat{s}_{r_{ij}}$ and $\hat{a}_{r_{ij}} \forall j \in \{1..d\}$\\
	{\bf $\triangleright$ Anomaly Score:}\\
	\ \ \ \ {\bf output} $score (r_{i},t) = \left(\hat{a}_{r_{i}} - \frac{\hat{s}_{r_{i}}}{t}\right)^2 \frac{t^2}{\hat{s}_{r_{i}}(t-1)} + \sum_{j=1}^{d} score(r_{ij},t)$\\
	}
\end{algorithm}

\subsection{Incorporating Correlation Between Features}
\label{sec:mstreamae}

In this section, we describe our \method-PCA, \method-IB and \method-AE approaches where we run the \method\ algorithm on a lower-dimensional embedding of the original data obtained using Principal Component Analysis (PCA) \cite{pearson1901liii}, Information Bottleneck (IB) \cite{tishby2000information} and Autoencoder (AE) \cite{hinton1994autoencoders} methods in a streaming manner.

Our motivation for combining PCA, IB and AE methods with \method\ is two-fold. Firstly, the low-dimensional representations learned by these algorithms incorporate correlation between different attributes of the record, making anomaly detection more effective. Secondly, a reduction in the dimensions would result in faster processing per record.

For all three methods, we first learn the dimensionality reduction transformation using a very small initial subset of $256$ records from the incoming stream. We then compute the embeddings for the subsequent records and pass them to \method\ to detect anomalies in an online manner.

\paragraph{\textbf{Principal Component Analysis}}

We choose PCA because it only requires one major parameter to tune: namely the dimension of the projection space. Moreover, this parameter can be set easily by analysis of the explained variance ratios of the principal components. Hence \method-PCA can be used as an off-the-shelf algorithm for streaming anomaly detection with dimensionality reduction.

\paragraph{\textbf{Information Bottleneck}}
Information bottleneck for dimensionality reduction can be posed as the following optimization problem:
$$
\min _{p(t | x)} I(X ; T)-\beta I(T ; Y)
$$
where $X$, $Y$, and $T$ are random variables. $T$ is the compressed representation of $X$, $I(X ; T)$ and $I(T ; Y)$ are the mutual information of $X$ and $T$, and of $T$ and $Y$, respectively, and $\beta$ is a Lagrange multiplier.
In our setting, $X$ denotes the multi-aspect data, $Y$ denotes whether the data is anomalous and $T$ denotes the dimensionally reduced features that we wish to find. Our implementation is based on the Neural Network approach for Nonlinear Information Bottleneck \cite{kolchinsky2017nonlinear}.

\paragraph{\textbf{Autoencoder}}
Autoencoder is a neural network based approach for dimensionality reduction. An autoencoder network consists of an encoder and a decoder. The encoder compresses the input into a lower-dimensional space, while the decoder reconstructs the input from the low-dimensional representation. Our experimental results in Section \ref{sec:experiment} show that even with a simple 3 layered autoencoder, \method-AE outperforms both \method-PCA and \method-IB.
 
\subsection{Time and Memory Complexity}
\label{sec:timecomplexity}
In terms of memory, \method\ only needs to maintain data structures over time, which requires memory proportional to $O(wbd)$, where $w$, $b$ and $d$ are the number of hash functions, the number of buckets in the data structures and the total number of dimensions; which is bounded with respect to the stream size.

For time complexity, the only relevant steps in Algorithm \ref{alg:mstream} are those that either update or query the data structures, which take $O(wd)$ (all other operations run in constant time). Thus, the time complexity per update step is $O(wd)$. 

\section{Experiments}
\label{sec:experiment}
In this section, we evaluate the performance of \method\ and \method-AE compared to \elliptic, LOF, I-Forest, \rcf\ and \densealert\  on multi-aspect data streams. We aim to answer the following questions:

\begin{enumerate}[label=\textbf{Q\arabic*.}]
\item {\bf Anomaly Detection Performance:} How accurately does \method\ detect real-world anomalies compared to baselines, as evaluated using the ground truth labels?
\item {\bf Scalability:} How does it scale with input stream length and number of dimensions? How does the time needed to process each input compare to baseline approaches?
\item {\bf Real-World Effectiveness:} Does it detect meaningful anomalies? Does it detect group anomalies?

\end{enumerate}

\paragraph{\textbf{Datasets}}

\emph{KDDCUP99} dataset \cite{KDDCup192:online} is based on the DARPA dataset and is among the most extensively used datasets for intrusion detection. Since the proportion of data belonging to the `attack' class is much larger than the proportion of data belonging to the `non-attack' class, we downsample the `attack' class to a proportion of $20\%$. \emph{KDDCUP99} has $42$ dimensions and $1.21$ million records.

\cite{ring2019survey} surveys different intrusion detection datasets and recommends to use the newer CICIDS \cite{sharafaldin2018toward} and UNSW-NB15 \cite{moustafa2015unsw} datasets. These contain modern-day attacks and follow the established guidelines for reliable intrusion detection datasets (in terms of realism, evaluation capabilities, total capture, completeness, and malicious activity) \cite{sharafaldin2018toward}.

\emph{CICIDS} 2018 dataset was generated at the Canadian Institute of Cybersecurity. Each record is a flow containing features such as Source IP Address, Source Port, Destination IP Address, Bytes, and Packets. These flows were captured from a real-time simulation of normal network traffic and synthetic attack simulators. This consists of the \emph{CICIDS-DoS} dataset ($1.05$ million records, 80 features) and the \emph{CICIDS-DDoS} dataset ($7.9$ million records, 83 features). \emph{CICIDS-DoS} has $5\%$ anomalies whereas \emph{CICIDS-DDoS} has $7\%$ anomalies.

\emph{UNSW-NB 15} dataset was created by the Cyber Range Lab of the Australian Centre for Cyber Security (ACCS) for generating a hybrid of real modern normal activities and synthetic contemporary attack behaviors. This dataset has nine types of attacks, namely, Fuzzers, Analysis, Backdoors, DoS, Exploits, Generic, Reconnaissance, Shellcode and Worms. It has $49$ features and $2.5$ million records including $13\%$ anomalies.

\begin{table*}[!htb]
\centering
\caption{AUC of each method on different datasets.}
\label{tab:auc}
\begin{tabular}{@{}rccccccccc@{}}
\toprule
& Elliptic
 & LOF
 & I-Forest
 &  DAlert
 & RCF
 & \textbf{\method}
 & \textbf{\method-PCA}
 & \textbf{\method-IB}
 & \textbf{\method-AE} \\ \midrule
 \textbf{KDD} & $0.34 \pm 0.025$ & $0.34$ &  $0.81 \pm 0.018$ & $0.92$ & $0.63$ &   $0.91 \pm 0.016$ & $0.92 \pm 0.000$ & $\mathbf{0.96} \pm 0.002$ & $\mathbf{0.96} \pm 0.005$ \\
 \textbf{DoS} & $0.75 \pm 0.021$ & $0.50$ & $0.73 \pm 0.008$ & $0.61$  & $0.83$ & $0.93 \pm 0.001$ & $0.92 \pm 0.001$ & $\mathbf{0.95} \pm 0.003$ & $0.94 \pm 0.001$ \\
  \textbf{UNSW} & $0.25 \pm 0.003$ & $0.49$ & $0.84 \pm 0.023$ & $0.80$ & $0.45$ & $0.86 \pm 0.001$ & $0.81 \pm 0.001$ & $0.82 \pm 0.001$ & $\mathbf{0.90} \pm 0.001$ \\
\textbf{DDoS} & $0.57 \pm 0.106$ & $0.46$ & $0.56 \pm 0.021$ & $--$ & $0.63$ &  $0.91 \pm 0.000$ & $\mathbf{0.94} \pm 0.000$ & $0.82 \pm 0.000$ & $0.93 \pm 0.000$ \\
\bottomrule
\end{tabular}
\end{table*}

\paragraph{\bf Baselines}
As described in Section \ref{sec:rel}, we are in the streaming unsupervised learning regime, and hence do not compare with supervised or offline algorithms.

We consider \elliptic, \lof, \iso, STA, MASTA, STenSr, \densealert\ and \rcf\ since they operate on multi-aspect data, however, due to a large number of dimensions, even sparse tensor versions of STA/MASTA/STenSr run out of memory on these datasets. So, we compare with \elliptic, \lof, \iso, \densealert\ and \rcf.

\paragraph{\bf Evaluation Metrics}
All the methods output an anomaly score per edge (higher is more anomalous). We plot the ROC curve, which compares the True Positive Rate (TPR) and False Positive Rate (FPR), without needing to fix any threshold. We also report the ROC-AUC (Area under the ROC curve).

\paragraph{\bf Experimental Setup}
All experiments are carried out on a $2.4 GHz$ Intel Core $i9$ processor, $32 GB$ RAM, running OS $X$ $10.15.2$. We implement \method\ in C\texttt{++}. We use $2$ independent copies of each hash function, and we set the number of buckets to 1024. We set the temporal decay factor $\alpha$ as $0.85$ for \emph{KDDCUP99}, $0.95$ for \emph{CICIDS-DoS} and \emph{CICIDS-DDoS}, and $0.4$ for \emph{UNSW-NB 15} due to its higher time granularity. Note that \method\ is not sensitive to variation of $\alpha$ parameter as shown in Table \ref{tab:FactorVsAUC}. Since \emph{KDDCUP99} dataset does not have timestamps, we apply the temporal decay factor once every 1000 records. We discuss the influence of temporal decay factor $\alpha$ on the ROC-AUC in Appendix \ref{app:3}.

To demonstrate the robustness of our proposed approach, we set the output dimension of \method-PCA, \method-IB and \method-AE for all datasets to a common value of $12$ instead of searching individually on each method and dataset. We reduce the real-valued columns to $12$ dimensions and then pass these along with the categorical columns to \method. Results on varying the number of output dimensions can be found in the Appendix. For \method-PCA we use the open-source implementation of PCA available in the scikit-learn \cite{scikit-learn} library. Parameters for \method-AE and \method-IB are described in Appendix \ref{sec:app1}.

We use open-sourced implementations of \densealert\ and \rcf, provided by the authors, following parameter settings as suggested in the original papers. For \elliptic, \lof\ and \iso\ we use the open-source implementation available in the scikit-learn \cite{scikit-learn} library. We also pass the true anomaly percentage to \elliptic, \lof\ and \iso\ methods, while the remainder of the methods do not require the anomaly percentage.

All the experiments, unless explicitly specified, are performed $5$ times for each parameter group, and the mean and standard deviation values are reported.

\subsection{Anomaly Detection Performance}

Figure \ref{fig:ROC} plots the ROC curve for \method, \method-PCA, \method-IB\ and \method-AE along with the baselines, \elliptic, \lof, \iso, \densealert\ and \rcf\ on \emph{CICIDS-DoS} dataset. 
We see that \method, \method-PCA, \method-IB\ and \method-AE achieve a much higher ROC-AUC ($0.92-0.95$) compared to the baselines. \method\ and its variants achieve at least $50\%$ higher AUC than \densealert, $11\%$ higher than \rcf\ $26\%$ higher than \iso, $23\%$ higher than \elliptic\ and  $84\%$ higher than \lof.

\begin{figure}[!htb]
        \center{\includegraphics[width=\columnwidth]
        {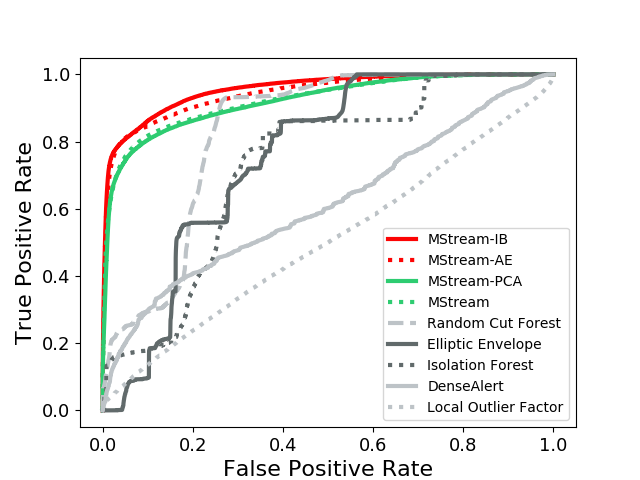}}
        \caption{\label{fig:ROC} ROC on \emph{CICIDS-DoS} dataset.}
        \Description{\label{fig:ROC} ROC on \emph{CICIDS-DoS} dataset.}
\end{figure}

Table \ref{tab:auc} shows the AUC of \elliptic, \lof, \iso, \densealert, \rcf\ and \method\ on \emph{KDDCUP99}, \emph{CICIDS-DoS}, \emph{UNSW-NB 15} and \emph{CICIDS-DDoS} datasets. We report a single value for \lof\ and \densealert\ because these are non-randomized methods. We also report a single value for \rcf\ because we use the parameters and random seed of the original implementation. \densealert\ performs well on small sized datasets such as \emph{KDDCUP99} but as the dimensions increase, its performance decreases. On the large \emph{CICIDS-DDoS} dataset \densealert\ runs out of memory. We observe that \method\ outperforms all baselines on all datasets. By learning the correlation between features, \method-AE achieves higher ROC-AUC than \method, and performs comparably or better than \method-PCA and \method-IB. We also discuss evaluating the ROC-AUC in a streaming manner in Appendix \ref{app:4}.

Figure \ref{fig:AUC} plots ROC-AUC vs. running time (log-scale, in seconds, excluding I/O) for the different methods on the \emph{CICIDS-DoS} dataset. We see that \method, \method-PCA, \method-IB and \method-AE achieve $11\%$ to $90\%$ higher AUC compared to baselines, while also running almost two orders of magnitude faster.

\begin{figure}[!htb]
        \center{\includegraphics[width=\columnwidth]
        {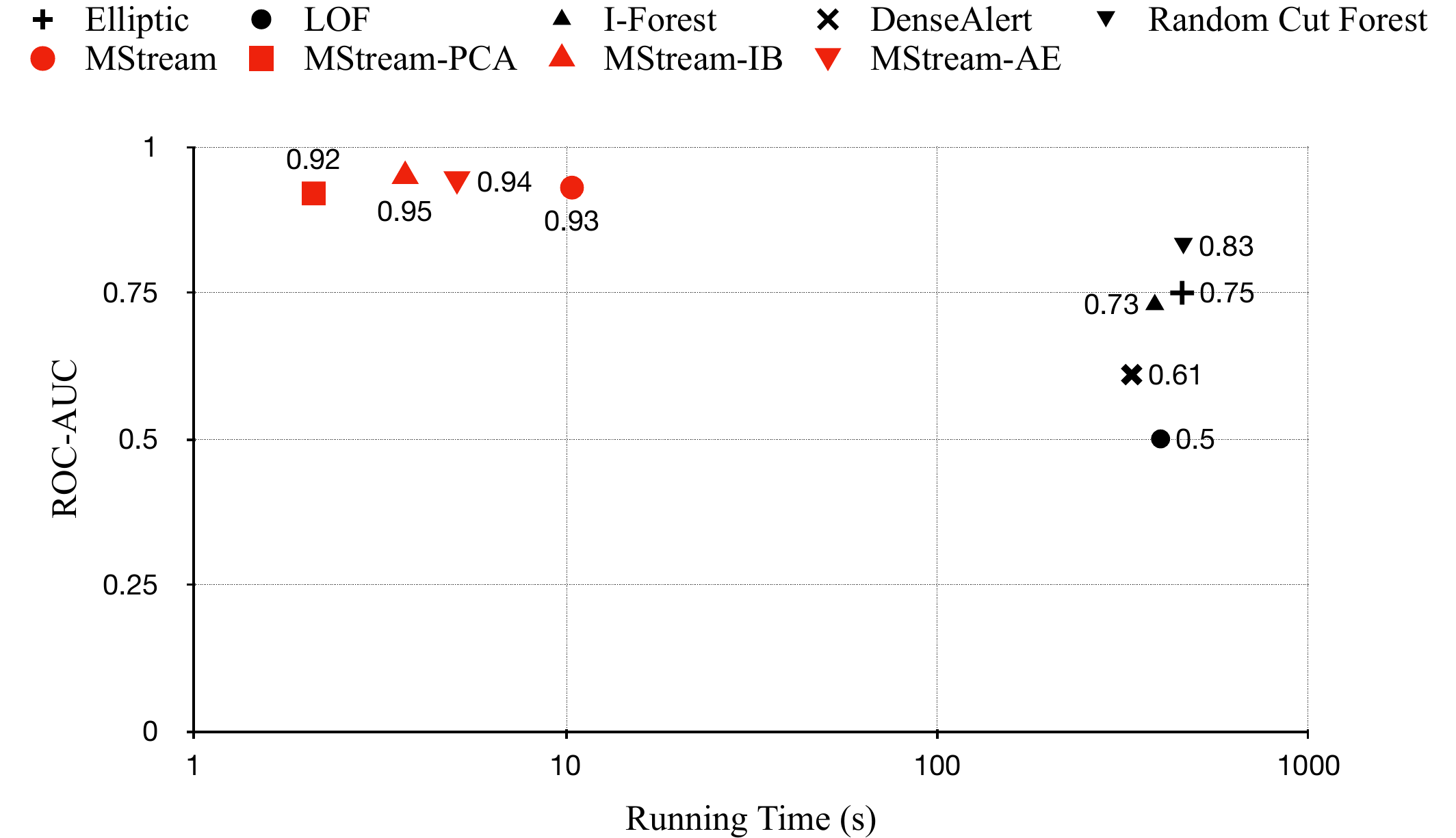}}
        \caption{\label{fig:AUC} ROC-AUC vs time on \emph{CICIDS-DoS} dataset.}
        \Description{\label{fig:AUC} ROC-AUC vs time on \emph{CICIDS-DoS} dataset.}
\end{figure}

\subsection{Scalability}

\begin{table*}[!htb]
\centering
\caption{Running time of each method on different datasets in seconds.}
\label{tab:times}
\begin{tabular}{@{}rccccccccc@{}}
\toprule
& Elliptic
 & LOF
 & I-Forest
 &  DAlert
 & RCF
 & \textbf{\method}
 & \textbf{\method-PCA}
 & \textbf{\method-IB}
 & \textbf{\method-AE} \\ \midrule
 \textbf{KDD} & $216.3$ & $1478.8$ &  $230.4$ & $341.8$ & $181.6$ & $4.3$ & ${2.5}$ & ${3.1}$ & ${3.1}$ \\
 \textbf{DoS} & $455.8$ & $398.8$ & $384.8$ & $333.4$  & $459.4$ & $10.4$ & ${2.1}$ & $3.7$ & $5.1$ \\
 \textbf{UNSW} & $654.6$ & $2091.1$ & $627.4$ & $329.6$ & $683.8$ & $12.8$ & ${6.6}$ & $8$ & $8$ \\
\textbf{DDoS} & $3371.4$ & $15577s$ & $3295.8$ & $--$ & $4168.8$ & $61.6$ & ${16.9}$ & $25.6$ & $27.7$ \\
\bottomrule
\end{tabular}
\end{table*}

Table \ref{tab:times} shows the time it takes \elliptic, \lof, \iso, \densealert, \rcf, \method\ and \method-AE to run on \emph{KDDCUP99}, \emph{CICIDS-DoS}, \emph{UNSW-NB 15} and \emph{CICIDS-DDoS} datasets. We see that \method\ runs much faster than the baselines: for example, \method\ is $79$ times faster than \densealert\ on the \emph{KDDCUP99} dataset. \method-PCA, \method-IB and \method-AE incorporate dimensionality reduction and are therefore faster than \method: for example, \method-AE is $1.38$ times faster than \method\ and $110$ times faster than \densealert\ on the \emph{KDDCUP99} dataset.

Figure \ref{fig:scaling1} shows the scalability of \method\ with respect to the number of records in the stream (log-scale). We plot the time needed to run on the (chronologically) first $2^{12}, 2^{13}, 2^{14},...,2^{20}$ records of the \emph{CICIDS-DoS} dataset. Each record has $80$ dimensions. This confirms the linear scalability of \method\ with respect to the number of records in the input stream due to its constant processing time per record.

\begin{figure}[!htb]
        \center{\includegraphics[width=\columnwidth]
        {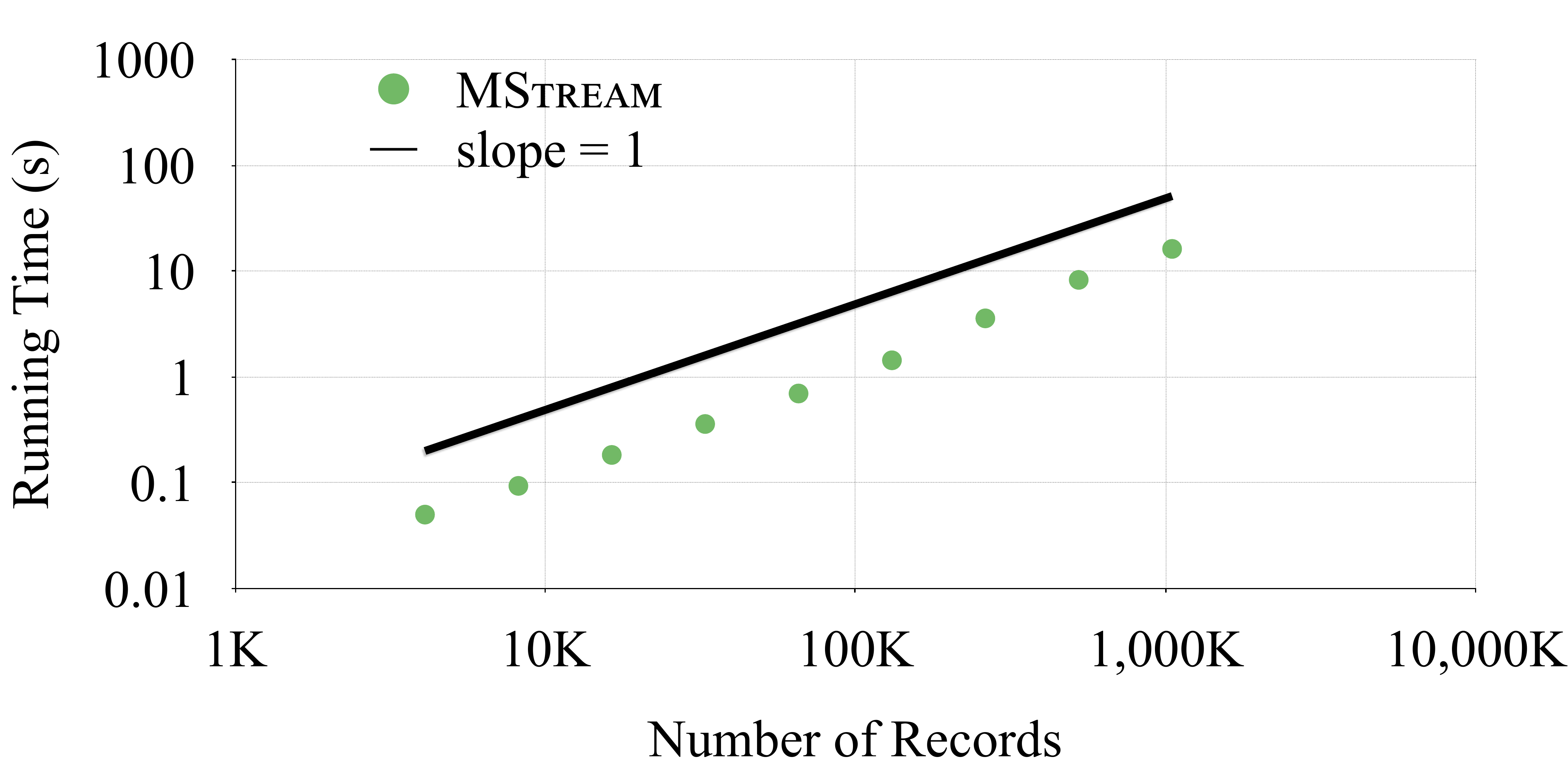}}
        \caption{\label{fig:scaling1} \method\ scales linearly with the number of records in \emph{CICIDS-DoS}.}
        \Description{\label{fig:scaling1} \method\ scales linearly with the number of records in \emph{CICIDS-DoS}.}
\end{figure}

Figure \ref{fig:scaling2} shows the scalability of \method\ with respect to the number of dimensions (linear-scale). We plot the time needed to run on the first $10, 20, 30,...,80$ dimensions of the \emph{CICIDS-DoS} dataset. This confirms the linear scalability of \method\ with respect to the number of dimensions in the input data.

\begin{figure}[!htb]
        \center{\includegraphics[width=\columnwidth]
        {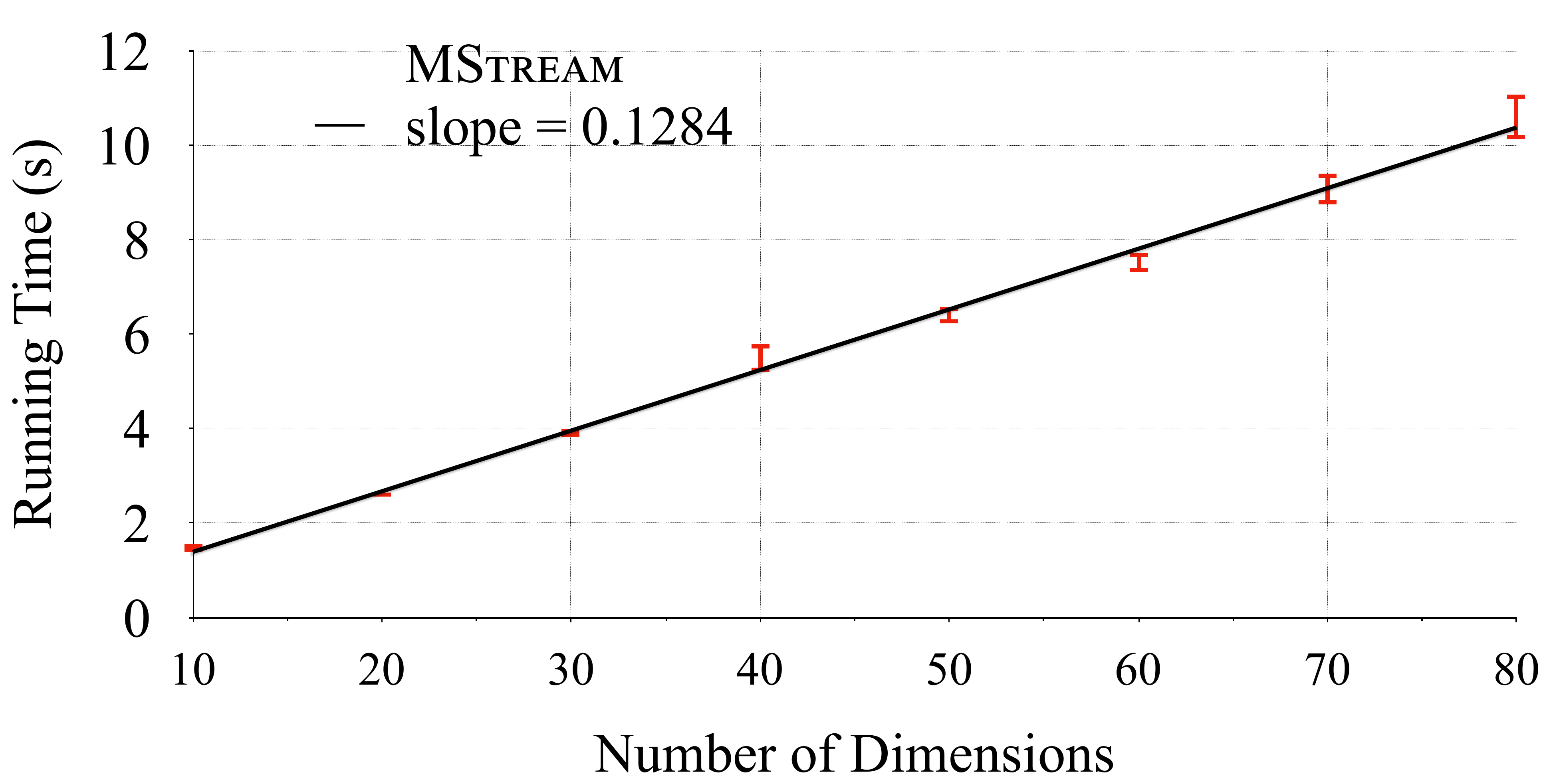}}
        \caption{\label{fig:scaling2} \method\ scales linearly with the number of dimensions in \emph{CICIDS-DoS}.}
        \Description{\label{fig:scaling2} \method\ scales linearly with the number of dimensions in \emph{CICIDS-DoS}.}
\end{figure}

Figure \ref{fig:rows} shows the scalability of \method\ with respect to the number of hash functions (linear-scale). We plot the time taken to run on the \emph{CICIDS-DoS} dataset with $2, 3, 4$ hash functions. This confirms the linear scalability of \method\ with respect to the number of hash functions.

\begin{figure}[!htb]
        \center{\includegraphics[width=\columnwidth]
        {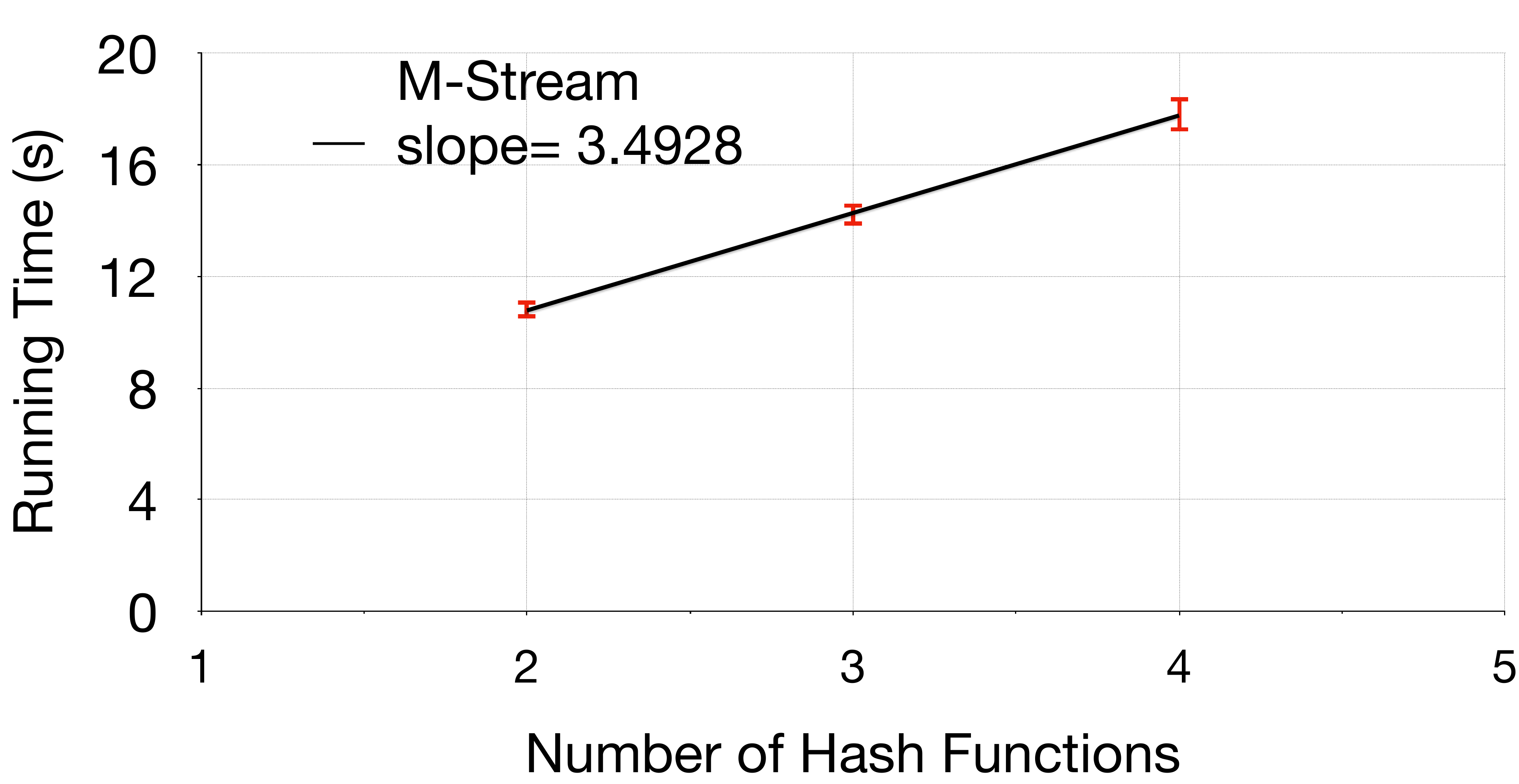}}
        \caption{\label{fig:rows} \method\ scales linearly with the number of hash functions in \emph{CICIDS-DoS}.}
        \Description{\label{fig:rows} \method\ scales linearly with the number of hash functions in \emph{CICIDS-DoS}.}
\end{figure}

Since \method-PCA, \method-IB\ and \method-AE apply \method\ on the lower-dimensional features obtained using an autoencoder, they are also scalable.

Figure \ref{fig:frequency} plots a frequency distribution of the time taken (in microseconds) to process each record in the \emph{CICIDS-DoS} dataset. \method\ processes $957K$ records within $10\mu s$ each, $60K$ records within $100\mu s$ each and remaining $30K$ records within $1000\mu s$ each.

\begin{figure}[!htb]
        \center{\includegraphics[width=\columnwidth]
        {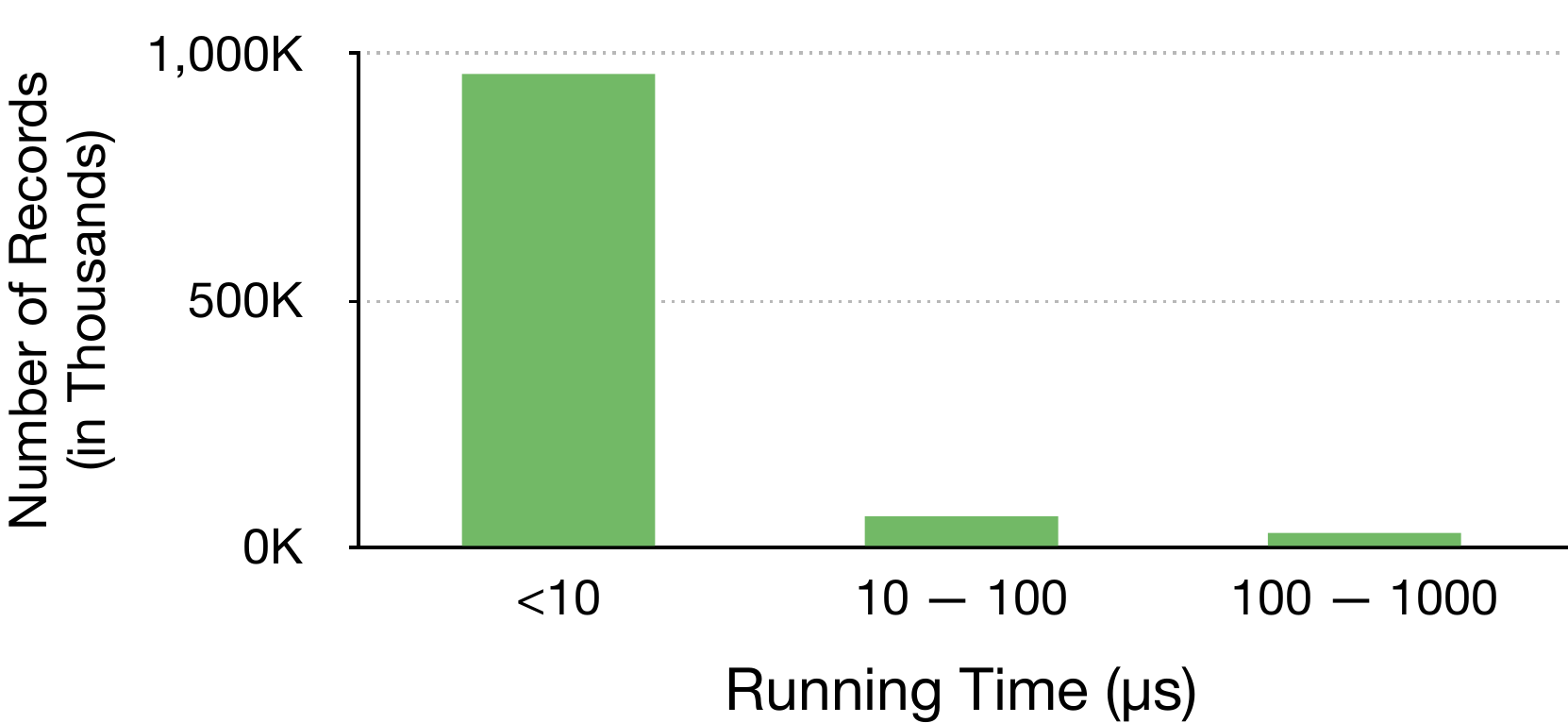}}
        \caption{\label{fig:frequency} Distribution of processing times for $\sim1.05M$ records of the \emph{CICIDS-DoS} dataset.}
        \Description{\label{fig:frequency} Distribution of processing times for $\sim1.05M$ records of the \emph{CICIDS-DoS} dataset.}
\end{figure}

\subsection{Discoveries}
We plot normalized anomaly scores over time using \elliptic, \lof, \iso, \densealert, \rcf\ and \method\ on the \emph{CICIDS-DoS} dataset in Figure \ref{fig:dos}. To visualize, we aggregate records occurring in each minute by taking the max anomaly score per minute, for a total of $565$ minutes. Ground truth values are indicated by points plotted at $y=0$ (i.e. normal) or $y=1$ (anomaly).

\lof\ and \densealert\ miss many anomalies whereas \elliptic, \iso\ and \rcf\ output many high scores unrelated to any attacks. This is also reflected in Table \ref{tab:auc} and shows that \method\ is effective in catching real-world anomalies.

\begin{figure}[!htb]
        \center{\includegraphics[width=\columnwidth]
        {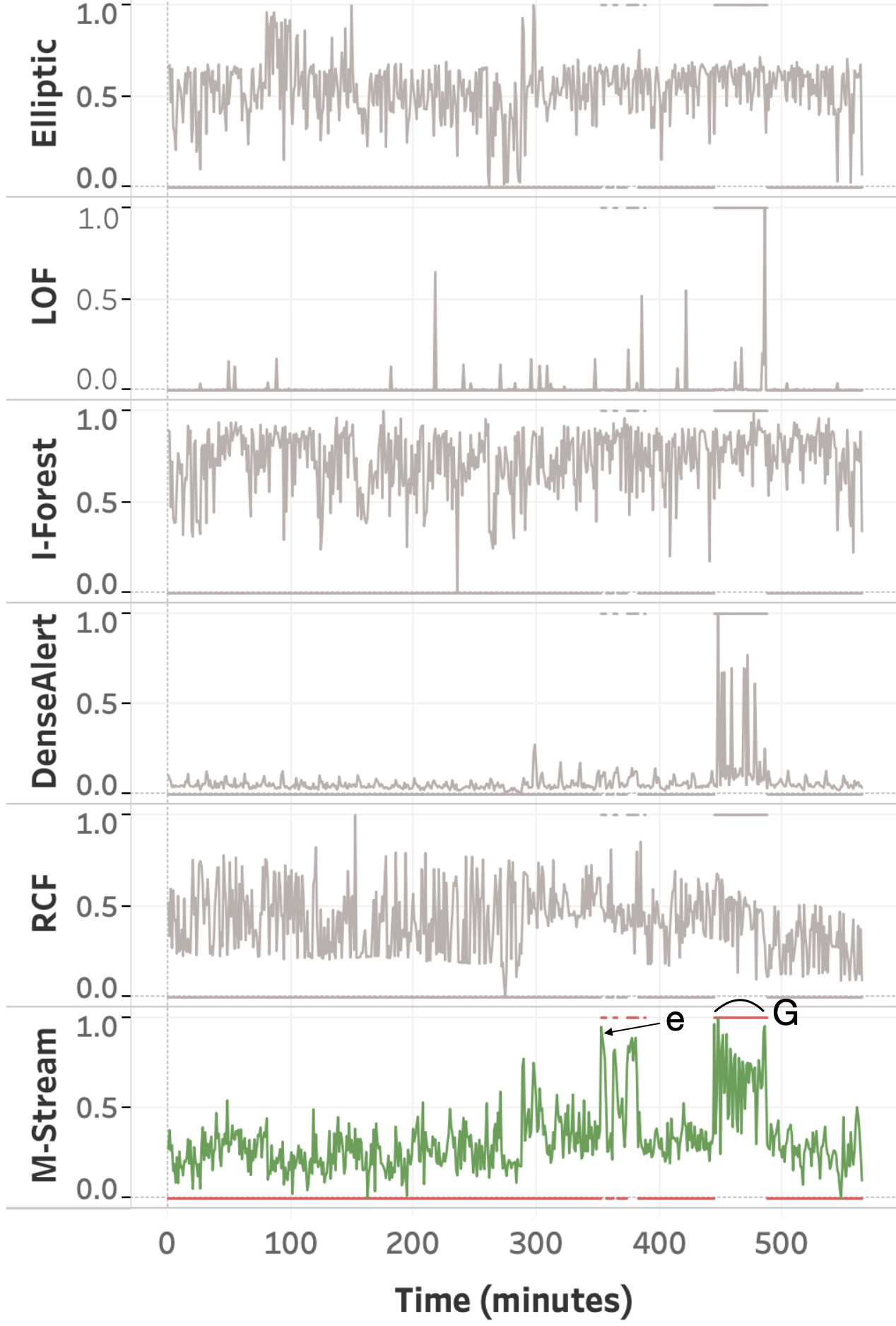}}
        \caption{\label{fig:dos} Plots of anomaly scores over time; spikes for \method\ correspond to the ground truth events in \emph{CICIDS-DoS}, but not for baselines.}
        \Description{\label{fig:dos} Plots of anomaly scores over time; spikes for \method\ correspond to the ground truth events in \emph{CICIDS-DoS}, but not for baselines.}
\end{figure}

\textbf{Group anomaly detection:}
In Figure \ref{fig:dos}, $G$ is a group anomaly which \method\ is able to detect, whereas \elliptic, \lof\ and \iso\ completely miss it. \densealert\ and \rcf\ partially catch it, but are also not fully effective in such high-dimensional datasets. This shows the effectiveness of \method\ in catching group anomalies such as DoS and DDoS attacks.

\textbf{Explainability:}
As \method\ estimates feature-specific anomaly scores before aggregating them, it is interpretable. For a given anomaly, we can rank the features according to their anomaly scores. We can then explain which features were most responsible for the anomalousness of a record in an unsupervised setting.

For example, in Figure \ref{fig:dos}, \method\ finds that $e$ is an anomaly that occurs due to the Flow IAT Min feature. This agrees with \cite{sharafaldin2018toward}, which finds that the best feature set for DoS using a Random Forest approach (supervised learning; in contrast, our approach does not require labels) are B.Packet Len Std, Flow IAT Min, Fwd IAT Min, and Flow IAT Mean.

\section{Conclusion}
In this paper, we proposed \method\ for detecting group anomalies in multi-aspect streams, and \method-PCA, \method-IB\ and \method-AE\ which incorporate dimensionality reduction to improve accuracy and speed. Future work could consider more complex combinations (e.g. weighted sums) of anomaly scores for individual attributes. Our contributions are:
\begin{enumerate}
    \item {\bf Multi-Aspect Group Anomaly Detection:} We propose a novel approach for detecting group anomalies in multi-aspect data, including both categorical and numeric attributes.
    \item {\bf Streaming Approach:} Our approach processes the data in a fast and streaming fashion, performing each update in constant time and memory.
    \item {\bf Effectiveness:} Our experimental results show that \method\ outperforms baseline approaches.
\end{enumerate}



\begin{acks}
This work was supported in part by \grantsponsor{odprt}{NUS ODPRT}{} Grant \grantnum{odprt}{R-252-000-A81-133}. The authors would like to thank Thijs Laarhoven for his suggestions.
\end{acks}

\bibliographystyle{ACM-Reference-Format}
\bibliography{references}

\appendix

\section{Influence of temporal decay factor}
\label{app:3}
Table~\ref{tab:FactorVsAUC} shows the influence of the temporal decay factor $\alpha$ on the ROC-AUC for \method\ on \emph{CICIDS-DoS} dataset. We see that $\alpha=0.95$ gives the maximum ROC-AUC for \method\ ($0.9326 \pm 0.0006$), as also shown in Table~\ref{tab:auc}.

\begin{table}[!htb]
		\centering
		\caption{Influence of temporal decay factor $\alpha$ on the
	    ROC-AUC in \method\  on \emph{CICIDS-DoS} dataset.}
		\label{tab:FactorVsAUC}
		\begin{tabular}{@{}ccc@{}}
			\toprule
			$\alpha$ & ROC-AUC \\
			\midrule
		$0.1$ & $0.9129 \pm 0.0004$ \\
        $0.2$ & $0.9142 \pm 0.0009$\\
        $0.3$ & $0.9156	\pm 0.0006$ \\
        $0.4$ & $0.9164	\pm 0.0014$ \\
        $0.5$ & $0.9163	\pm 0.0005$ \\
        $0.6$ & $0.917 \pm 0.0005$ \\
        $0.7$ & $0.9196	\pm 0.0015$ \\
        $0.8$ & $0.9235	\pm 0.0003$ \\
        $0.9$ & $0.929 \pm	0.0003$ \\
        $0.95$ & $0.9326 \pm 0.0006$ \\
			\bottomrule
		\end{tabular}
		
\end{table}

\section{Influence of dimensions}
\label{app:5}
Table~\ref{tab:dimensions} shows the influence of the output dimensions on the ROC-AUC for \method-PCA, \method-IB, and \method-AE \emph{KDDCUP99} dataset. We see that all methods are robust to the variation in output dimensions.

\begin{table}[H]
\centering
\caption{Influence of Output Dimensions on the ROC-AUC of \method-PCA, \method-IB, and \method-AE on \emph{KDDCUP99} dataset.}
\label{tab:dimensions}
\resizebox{\columnwidth}{!}{
\begin{tabular}{@{}rcccc@{}}
\toprule
\textbf{Dimensions}
 & \textbf{\method-PCA}
 & \textbf{\method-IB}
 & \textbf{\method-AE} \\ \midrule
 $4$ & $0.93$ & $0.95$ & $0.95$\\
 $8$ & $0.94$ & $0.95$ & $0.93$ \\
$12$ & $0.92$ & $0.96$ & $0.96$ \\
 $16$ & $0.87$ & $0.96$ & $0.96$ \\
\bottomrule
\end{tabular}}
\end{table}

\section{Dimensionality Reduction}
\label{sec:app1}
For \method-IB, we used an online implementation, \url{https://github.com/burklight/nonlinear-IB-PyTorch} for the underlying Information Bottleneck algorithm with $\beta=0.5$ and the variance parameter set to a constant value of $1$. The network was implemented as a $2$ layer binary classifier. For \method-AE, the encoder and decoder were implemented as single layers with ReLU activation.

Table \ref{tab:aearch} shows the network architecture of the autoencoder. Here $n$ denotes the batch size, and $d$ denotes the input data dimensions. The input data dimensions for each dataset are described in Section \ref{sec:experiment}.  
\begin{table}[!htb]
\begin{center}
\caption{Autoencoder Architecture}
\label{tab:aearch}
\begin{tabular}{@{}rccc@{}}
\toprule
\textbf{Index} & \textbf{Layer} & \textbf{Output Size}  \\ \midrule
$1$ \ \ \ \ & Linear & $n\times12$\\
$2$ \ \ \ \ & ReLU & $n\times12$\\
$3$ \ \ \ \ & Linear & $n\times d$\\
\bottomrule
\end{tabular}
\end{center}
\end{table}

We used Adam Optimizer to train both these networks with $\beta_1 = 0.9$ and $\beta_2 = 0.999$. Grid Search was used for hyperparameter tuning:  Learning Rate was searched on $[1\mathrm{e}-2, 1\mathrm{e}-3, 1\mathrm{e}-4, 1\mathrm{e}-5]$, and number of epochs was searched on $[100, 200, 500, 1000]$. The final values for these  can be found in Table \ref{tab:ibparam}.
\begin{table}[!htb]
\centering
\caption{\method-IB parameters for different datasets.}
\label{tab:ibparam}
\begin{tabular}{@{}rcccc@{}}
\toprule
& \multicolumn{2}{c}{\method-IB} & \multicolumn{2}{c}{\method-AE} \\
 \textbf{Dataset} & Learning Rate  &  Epochs & Learning Rate  & Epochs  \\ \midrule
 \textbf{KDD} & $1\mathrm{e}-2$ & $100$ & $1\mathrm{e}-2$ & $100$\\
 \textbf{DoS} & $1\mathrm{e}-5$ &  $200$ & $1\mathrm{e}-2$ &  $1000$\\
 \textbf{UNSW} & $1\mathrm{e}-2$ & $100$ & $1\mathrm{e}-2$ & $100$\\
\textbf{DDoS} & $1\mathrm{e}-3$ & $200$ & $1\mathrm{e}-3$ & $100$\\
\bottomrule
\end{tabular}
\end{table}

\section{Evaluating ROC-AUC in a streaming manner}
\label{app:4}
Table~\ref{tab:streamingauc} shows the ROC-AUC for \method-AE on \emph{KDDCUP99} when evaluated over the stream. The evaluation is done on all records seen so far and is performed after every $100K$ records. We see that as the stream length increases, ROC-AUC for \method-AE converges to $0.96$, as also shown in Table~\ref{tab:auc}.

\begin{table}[H]
\centering
\caption{Evaluating ROC-AUC of \method-AE in a streaming manner on \emph{KDDCUP99} dataset.}
\label{tab:streamingauc}
\begin{tabular}{@{}rcc@{}}
\toprule
	Stream Size & ROC-AUC \\ \midrule
 $100K$ & $0.912488$ \\
        $200K$ & $0.895391$ \\
        $300K$ & $0.855598$ \\
        $400K$ & $0.934532$ \\
        $500K$ & $0.965250$ \\
        $600K$ & $0.953906$ \\
        $700K$ & $0.947531$ \\
        $800K$ & $0.961340$ \\
        $900K$ & $0.973217$ \\
        $1000K$ & $0.970212$ \\
        $1100K$ & $0.967215$ \\
        $1200K$ & $0.959664$ \\
\bottomrule
\end{tabular}
\end{table}

\end{document}